\documentclass[letterpaper]{article}
\usepackage[preprint]{aaai2027}
\usepackage[hyphens]{url}
\usepackage{graphicx}
\usepackage{natbib}
\usepackage{caption}
\usepackage{booktabs}
\usepackage{amsmath}
\usepackage{amssymb}
\usepackage{xspace}
\newcommand{\method}{IAD-Unify\xspace}
\newcommand{\dataset}{Anomaly Evidence\xspace}
\newcommand{\R}{\mathbb{R}}
\newcommand{\xobs}{x^{\mathrm{obs}}}

\newcommand{\xsrc}{x^{\mathrm{src}}}
\newcommand{\mseg}{M_{\mathrm{seg}}}
\newcommand{\mpred}{M_{\mathrm{pred}}}
\newcommand{\medit}{M_{\mathrm{edit}}}
\newcommand{\Efield}{E_{\mathrm{dense}}}

\title{IAD-Unify: Task-Specific Interfaces for Industrial Anomaly Understanding, Segmentation, and Generation}
\author{Haoyu Zheng$^1$, Jiang Liu$^1$, Jiaqi Zhu$^2$, Feifei Shao$^1$, Zheqi Lv$^3$, Wenqiao Zhang$^1$\corresponding}
\affiliations{
$^1$Zhejiang University, China\\
$^2$National University of Singapore, Singapore\\
$^3$Cornell University, USA
}

\begin{document}
\maketitle

\begin{abstract}
Industrial anomaly inspection requires complementary capabilities: explaining
an observed defect, localizing its pixels, and synthesizing a controlled edit.
We present \method, a unified architecture connecting a
multimodal language model (MLLM), dense visual expert, and diffusion editor
through task-specific token interfaces. A multi-reference DINOv2 pathway forms a
dense anomaly field and compresses its 1,369 cells into 81 structured
evidence tokens. Qwen3.5 consumes these tokens for grounded answers and, with
32 task tokens, converts them into a semantic residual over the dense mask.
A separate 256-query interface resamples Qwen states
into Stable Diffusion's complete cross-attention context, while the editor
retains its source latent and hard-mask inputs. The
task pathways share one Qwen adaptation without forcing every task through the same
visual bottleneck; in particular, evidence tokens are excluded from the
generation pathway. Staged optimization initializes dense evidence, aligns it
with language, calibrates segmentation, and then pretrains and specializes the
diffusion editor while preserving earlier capabilities. We also construct
\dataset over
54,501 deduplicated industrial images. Its quality-controlled Anomaly Evidence
Compiler and Industrial Edit-Pair Compiler produce family-disjoint, validated
supervision through independent geometry, semantic-grounding, and
edit-consistency checks. This design
provides one fixed shared parameter set for
understanding, segmentation, and localized generation without conflating
their inputs, supervision, or outputs. The resulting model reaches
73.02\% MMAD Macro$_7$, the highest listed public segmentation AP average
(57.10\%) with one reference, and the lowest Controlled masked DINO distance
(0.3765), with complementary strengths across reasoning, pixel ranking, and
semantic edit fidelity.

\end{abstract}

\section{Introduction}
\label{sec:introduction}

Industrial anomaly inspection rarely ends with a binary decision. Once a
failure is observed, an inspection system must localize the affected region,
explain the visible evidence, and sometimes synthesize controlled variants for
training or process analysis. Existing methods address these needs separately:
anomaly detectors localize subtle defects
\citep{roth2022patchcore,batzner2024efficientad,damm2025anomalydino},
multimodal assistants interpret anomalous observations
\citep{gu2024anomalygpt,li2023myriad,jiang2026adcopilot}, and diffusion editors
generate defects inside specified regions
\citep{hu2024anomalydiffusion,zhuang2023powerpaint,ju2024brushnet}. Their
capabilities, however, remain split across models, input conventions, and
evaluation protocols.

\begin{figure*}[t]
  \centering
  \includegraphics[width=\textwidth]{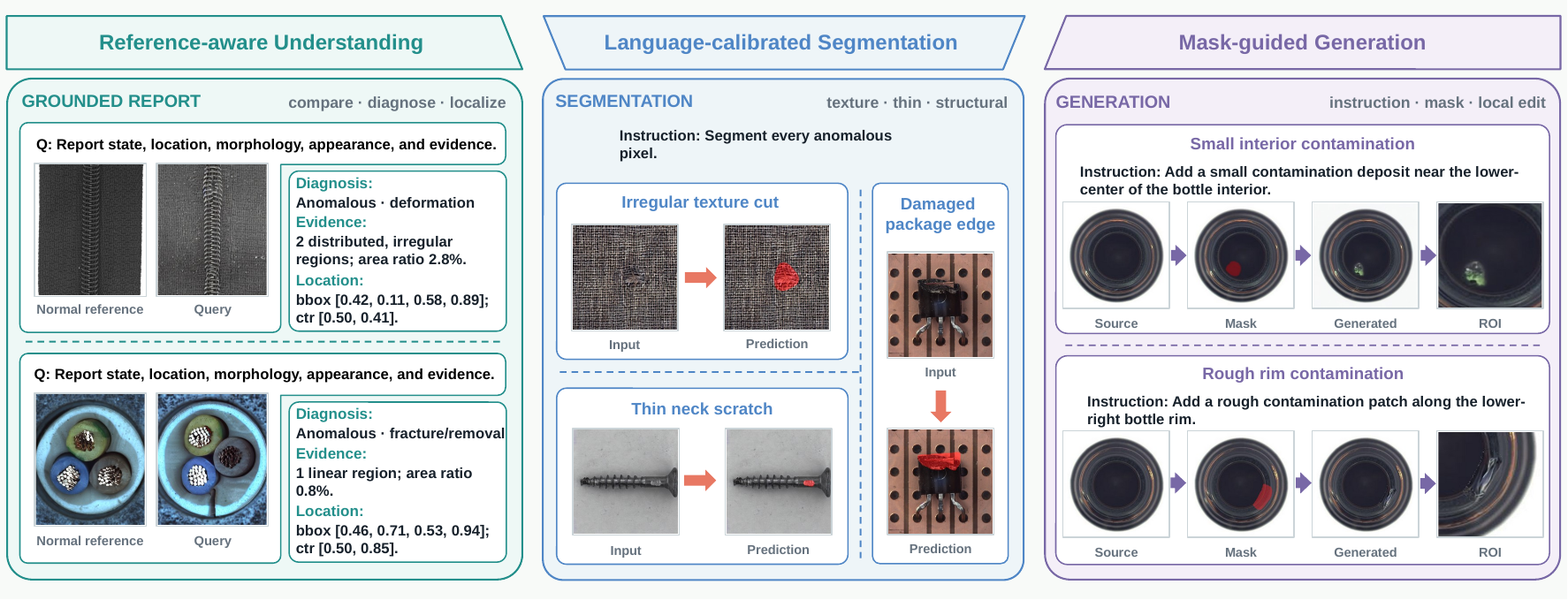}
  \caption{Capability overview of \method. Reference-aware understanding
  reports anomaly status and grounded evidence; language-calibrated segmentation
  predicts anomalous pixels; and mask-guided generation follows region-specific
  instructions over a real normal source. The generation examples share the
  same source while varying the edit instruction and mask.}
  \label{fig:capability-overview}
\end{figure*}

Open-domain systems increasingly unify visual understanding and generation with
shared or decoupled representations
\citep{xie2024showo,wu2025janus,wang2024emu3,wang2025ovisu1}. Their general
interfaces do not, however, preserve two forms of spatial conditioning central to industrial
inspection: normal-reference dense comparison for localization and
source-plus-hard-mask control for editing. Industrial unification must preserve
these task-specific inputs, not merely support both output modalities.

The same fragmentation appears in evaluation. MVTec AD, VisA, and MPDD provide
pixel annotations for localization
\citep{bergmann2019mvtec,zou2022spot,jezek2021mpdd}; MMAD evaluates multimodal
anomaly reasoning \citep{jiang2024mmad}; and editing benchmarks generally use
separate source--instruction pairs. The challenge is therefore to obtain one
model with a single set of parameters whose information flow remains valid across
incompatible task requirements, rather than placing three components behind one API. Scores from
these settings refer to different image populations and cannot by themselves
establish whether one model retains all three capabilities.

A uniform interface would discard task structure: localization needs dense
patch correspondence, language needs semantic and instructional context, and
inpainting needs a source latent, hard edit mask, and diffusion condition.
Forcing them through one bottleneck loses useful structure, whereas isolating
them forfeits parameter sharing. Our design principle is therefore
\emph{selective unification}: share parameters and evidence where their
semantics agree, but retain task-specific token interfaces and inputs where they do
not. Understanding and segmentation share observed-anomaly evidence;
generation follows a user instruction over a supplied source and edit region,
never a predicted anomaly mask.

We instantiate this principle in \method, one unified architecture for
industrial anomaly understanding, segmentation, and localized generation.
Figure~\ref{fig:capability-overview} previews the three user-facing pathways and
their distinct spatial inputs and outputs. A
DINOv2-L expert compares an observation with zero or more normal references and
produces a dense anomaly field \citep{oquab2023dinov2}. A structured connector
compresses its 1,369 spatial cells into 81 continuous Evidence Tokens for
Qwen3.5-9B \citep{qwen2026qwen35}. Qwen decodes grounded answers from this
evidence, while 32 Segmentation Tokens form a bounded semantic residual over
the dense mask. Generation takes a separate pathway: 256 learned Diffusion Queries read
Qwen's source- and instruction-conditioned states and become Stable Diffusion's
complete cross-attention context \citep{rombach2022ldm,pan2025metaquery}. This
query-conditioned pathway preserves the editor's nine-channel inpainting
input and excludes the 81 Evidence Tokens.

A dependency-aware optimization sequence gives each interface a defined
initialization point and stage-specific trainable parameter set. Each stage
inherits its predecessor and updates only that set, producing one model without
task-selected checkpoints or an unconstrained multi-task objective.

Cross-task training also requires consistent data roles. We introduce \dataset,
which compiles 54,501 deduplicated industrial images from MVTec AD, VisA, and MPDD
into family-disjoint, task-specific records. Its Anomaly Evidence Compiler establishes
duplicate-aware identity, partitions, and observable evidence; its Industrial
Edit-Pair Compiler creates spatially constrained editing records. Independent
geometry, semantic-grounding, and edit-consistency checks apply a common
validation standard without conflating segmentation labels, predictions, and
user edit masks.

The final model reaches 73.02\% Macro$_7$ on MMAD, the highest listed
public segmentation AP average (57.10\%) with one reference, and the lowest
Controlled masked DINO distance (0.3765). The resulting model offers a
competitive three-task performance profile with complementary strengths in
reasoning, pixel ranking, and semantic editing.

We summarize our main contributions as follows:
\begin{itemize}
  \item We formulate unified industrial anomaly inspection as one fixed
  architecture with selective parameter sharing and task-specific token interfaces for
  understanding, segmentation, and generation.
  \item We develop complementary evidence and editing pathways: an 81-token
  multi-reference evidence bottleneck with a 32-token semantic mask residual,
  and a separate 256-query interface that transfers source--instruction context
  to diffusion while preserving the inpainting model's original inputs.
  \item We introduce \dataset, a family-disjoint resource spanning 54,501
  deduplicated images, 24 products, and 104 defect variants, with validated
  supervision and field-level quality controls across all three tasks.
  \item We validate one set of final model parameters across benchmarked understanding,
  public one-reference segmentation, and matched localized editing, establishing
  a competitive three-task performance profile with complementary task strengths.
\end{itemize}

\section{Related Work}
\label{sec:related}

\paragraph{Industrial anomaly localization.}
Dense anomaly localization is dominated by specialist models trained from
normal data through patch-memory, distillation, and reconstruction
\citep{roth2022patchcore,batzner2024efficientad,deng2022rd4ad,zavrtanik2021draem}.
CutPaste and GLASS instead exploit artificial defects
\citep{li2021cutpaste,chen2024unified}, while WinCLIP, AnomalyCLIP, and
AnomalyDINO improve zero- or few-shot transfer
\citep{jeong2023winclip,zhou2026anomalyclip,damm2025anomalydino}.
UniAS unifies localization across categories \citep{ma2025towards}; we instead
use DINOv2 correspondence \citep{oquab2023dinov2} within a model that also
supports understanding and editing.

\paragraph{Multimodal anomaly understanding.}
Vision-language pretraining enables image-grounded dialogue
\citep{radford2021clip,liu2023visual}. Industrial systems ground this interface
with anomaly features: AnomalyGPT and Myriad connect visual experts to language
models \citep{gu2024anomalygpt,li2023myriad}, while
MAU-GPT, OmniAD, and AD-Copilot broaden the supported reasoning tasks
\citep{wang2026mau,zhao2025omniad,jiang2026adcopilot}. MMAD supplies a common
evaluation protocol \citep{jiang2024mmad}. Unlike these assistants, our model
also predicts dense masks and performs controlled edits.

\paragraph{Anomaly synthesis and localized editing.}
Diffusion and latent diffusion support text- and mask-conditioned generation
\citep{ho2020denoising,rombach2022ldm}. General editors follow instructions or
use masked-image conditioning
\citep{brooks2023instructpix2pix,zhuang2023powerpaint,ju2024brushnet}, whereas
AnomalyDiffusion and DualAnoDiff target industrial anomaly synthesis
\citep{hu2024anomalydiffusion,jin2025dual}. We instead study localized editing
from a source, hard mask, and defect instruction.

\paragraph{Unified understanding and generation.}
Show-o, Janus, Emu3, and Ovis-U1 combine visual understanding and generation
through unified tokens, decoupled encoders, or autoregressive image modeling
\citep{xie2024showo,wu2025janus,wang2024emu3,wang2025ovisu1}. These open-domain
systems do not directly encode industrial normal-reference comparison and
source-plus-mask editing. MetaQuery transfers MLLM states to diffusion through
learned queries \citep{pan2025metaquery}. Existing datasets separately emphasize
localization
\citep{bergmann2019mvtec,zou2022spot,jezek2021mpdd} or reasoning (MMAD)
\citep{jiang2024mmad}; \dataset instead
aligns family-disjoint records across all three tasks within one architecture.

\section{Method}
\label{sec:method}

\subsection{\dataset Compilation}
\label{sec:dataset}

\begin{figure*}[t]
  \centering
  \includegraphics[width=\textwidth]{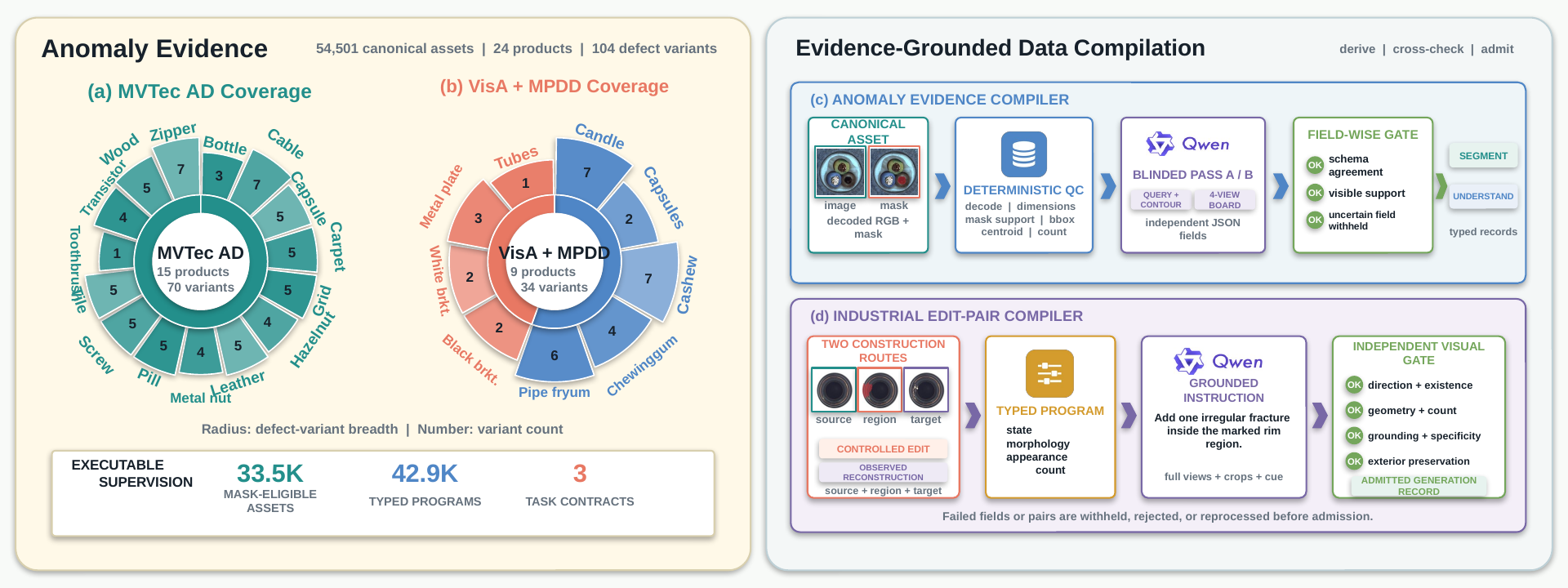}
  \caption{\dataset and its evidence-grounded compilation. Product-level bars
  summarize defect diversity across 24 products. The two compilers create
  verified records for understanding and segmentation, and grounded
  source--instruction--target records for generation.}
  \label{fig:dataset-compilation}
\end{figure*}

\dataset compiles MVTec AD \citep{bergmann2019mvtec}, VisA
\citep{zou2022spot}, and MPDD \citep{jezek2021mpdd} into validated,
task-specific records while preserving source provenance. Its \emph{Anomaly
Evidence Compiler} produces leakage-controlled supervision for segmentation
and understanding; its \emph{Industrial Edit-Pair Compiler} produces grounded,
spatially constrained records for localized generation.
Figure~\ref{fig:dataset-compilation} summarizes the resulting product coverage,
validated task records and the independent quality-control paths used by the two
compilers.

\paragraph{Duplicate-aware identity and partitioning.}
Exact decoded-RGB canonicalization reduces 59,916 image paths to 54,501 unique
images. Before constructing task examples, duplicate files, derived images,
reference-linked images, and samples from the same confirmed acquisition are
grouped into families; every question, mask, reference entry, and edit pair
inherits its family's partition. The split
contains 38,774 training, 4,663 development, and 9,535 test images; 1,529
ambiguous images are excluded by quality control. No duplicate-content,
derivation, or family relation crosses these partitions, which preserve upstream
test assignments.
This family-level assignment is applied before task records are constructed, so
expanding one image into questions, masks, or edit pairs cannot create a cross-split
shortcut.

\paragraph{Evidence quality control.}
Each image passes deterministic media, mask, and geometry checks. Geometric
fields are recomputed from versioned masks; semantic fields are proposed from
full-image and region views and independently checked against visible evidence.
Field-wise validation preserves separately verified anomaly status, localization, and
geometry supervision. Held-out understanding uses fixed task templates and
scoring rules,
and evaluation answers never enter compilation or model selection.
An uncertain appearance attribute therefore does not invalidate separately
verified anomaly-status or geometry fields; unsupported fields are withheld rather than
converted into supervision.

\paragraph{Task-specific supervision.}
The segmentation task predicts all visible anomalous pixels and uses
33,494 eligible images. The understanding task covers anomaly status,
localization, counting, visual characterization, and grounded reporting, with
33,801 eligible images and 42,941 structured task instances with schema-constrained answers
and task-specific scoring functions. Generation uses
$(\xsrc,\medit,t,x^{\mathrm{target}})$ tuples. We keep the dataset label
$\mseg$, model prediction $\mpred$, and user edit mask $\medit$ distinct;
generation never treats $\mpred$ as its required edit instruction.

\paragraph{Industrial Edit-Pair Compiler.}
The compiler instantiates structured defect specifications through two procedures under one
localized-editing setting. \emph{Controlled-edit} records alter a real normal
source only within $\medit$; \emph{masked-reconstruction} records pair an
observed anomaly with its region-masked input. Instructions are grounded in the
realized source--target change. Quality control requires visible, localized,
semantically supported, exterior-preserving edits; each accepted pair retains
its construction procedure and family-level split.
Checks apply to the realized pair rather than the intended defect program alone,
separating verified visual evidence from generation intent.

Together, the two compilers provide a consistent evidence layer for three
tasks while preserving their distinct inputs, labels, and evaluation
requirements.

\subsection{Unified Architecture and Task-Specific Interfaces}

\begin{figure*}[t]
  \centering
  \includegraphics[width=\textwidth]{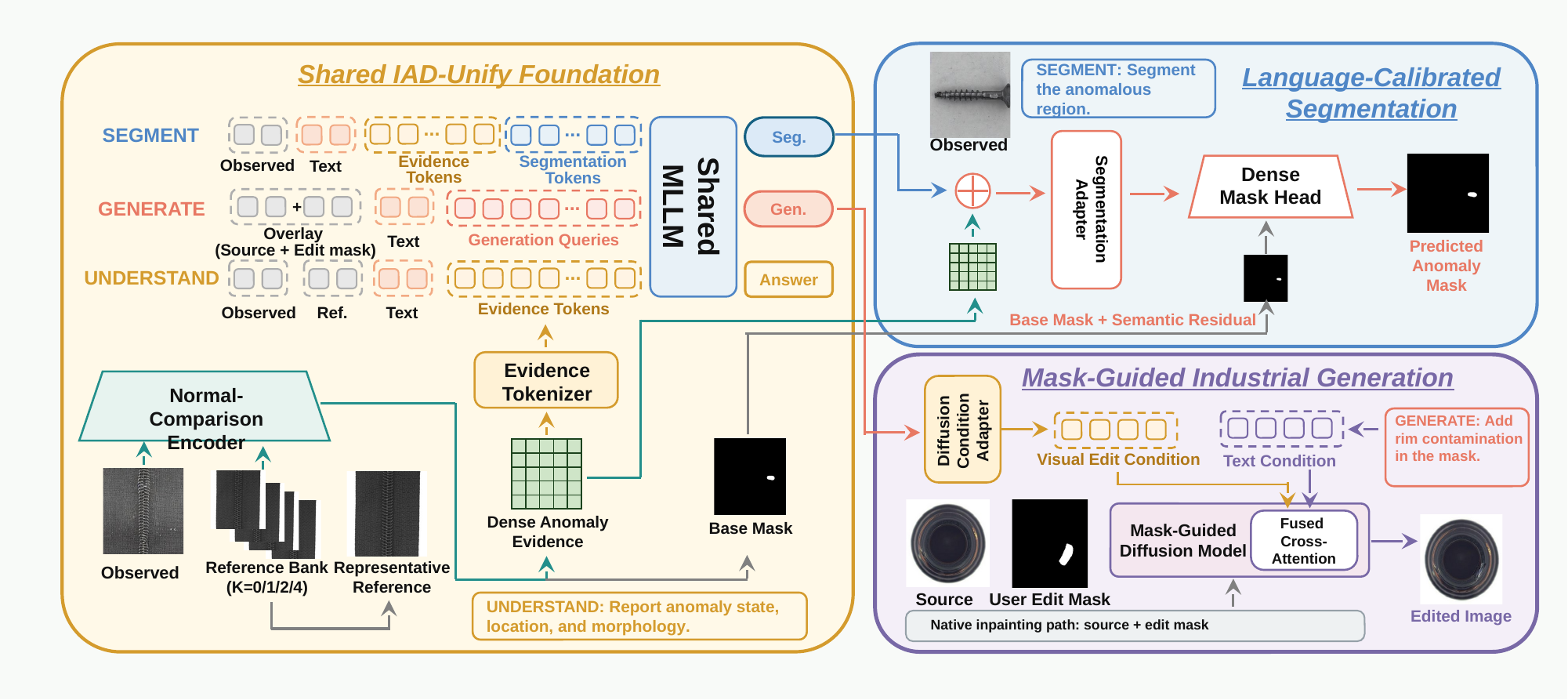}
  \caption{\textbf{Overview of \method.} A shared multimodal language model
  connects three task pathways while preserving their task-specific inputs and
  outputs. Understanding and segmentation consume dense anomaly evidence from
  an observed image and optional normal references. Generation instead uses a
  source image, a user edit mask, and an instruction; the predicted anomaly
  mask never serves as an edit command.}
  \label{fig:method-overview}
\end{figure*}

Figure~\ref{fig:method-overview} summarizes the unified architecture. Let $\xobs$
denote an observed image, $q$ a task instruction, and
$\mathcal{R}=\{x_k^{\mathrm{ref}}\}_{k=1}^{K}$ an optional normal-reference
bank, where $K\in\{0,1,2,4\}$. Understanding predicts answer tokens $y$, while
segmentation predicts $\mpred\in[0,1]^{H\times W}$. Generation instead receives
a source image $\xsrc$, instruction $t$, and user-supplied hard mask $\medit$
to produce $\hat{x}$. All pathways use the same model parameters.

The architecture combines DINOv2-L/14 \citep{oquab2023dinov2}, Qwen3.5-9B
\citep{qwen2026qwen35}, and Stable Diffusion 2 inpainting
\citep{rombach2022ldm} (SD2). Their foundation weights are largely frozen; adaptation
uses a shared Qwen LoRA \citep{hu2022lora}, explicit task tokens, task-specific
connectors, a dense evidence head, and selected diffusion blocks; all
other foundation parameters remain frozen.

Task-specific token interfaces separate shared representations by function. The dense field
$F$ anchors localization; Evidence Tokens expose it to language; Segmentation
Tokens write a bounded residual over the dense mask; and Diffusion Queries
condition diffusion on the source and instruction. Generation shares Qwen but
bypasses anomaly evidence, so a predicted region cannot silently become an edit
command.

Understanding and segmentation share the observed image and reference-conditioned
field, whereas generation separately encodes its source, instruction, and edit
mask. Parameters are therefore shared without conflating task-specific spatial
inputs and outputs.

\subsection{Multi-Reference Dense Evidence}

We extract multi-level DINO features from the observation and optional normal
references. Each query patch retrieves neighbors from the valid reference bank;
DINO uses the full bank, while Qwen sees at most one representative reference.
Explicit conditioning modes distinguish valid, missing, and query-only cases.

A shared evidence head combines query features, matched reference features,
correspondence uncertainty, spatial coordinates, and conditioning mode:
\begin{equation}
  (F,L_0)=\Efield(\xobs,\mathcal{R}),\qquad
  F\in\R^{256\times37\times37},
\end{equation}
where $L_0\in\R^{512\times512}$ gives the base-mask logits. Thus localization
remains dense before language-space compression.
Reference matching captures deviations from normal appearance, while query
features and validity encoding handle missing or mismatched references.

\paragraph{Structured evidence connector.}
The connector produces 81 continuous tokens: 64 evidence-weighted cells on an
$8\times8$ grid, 16 local-peak tokens, and one geometry token. A final
projection maps them to Qwen's embedding space,
\begin{equation}
  T_E=\operatorname{Conn}_{E}(F,L_0)\in\R^{81\times4096}.
\end{equation}
This mask-free representation preserves broad coverage, small or fragmented peaks,
and global extent without pooling an oracle mask. Its cardinality is independent
of reference count and anomaly area.

\subsection{Understanding and Segmentation}

\paragraph{Grounded understanding.}
Qwen receives the observation, optional representative reference, question,
and $T_E$, and is trained with assistant-token cross-entropy. During the
evidence-grounded language-alignment stage, raw Qwen image tokens are withheld
so that answers must pass through the evidence bottleneck; the full inference
model restores Qwen's original visual input.
This intervention makes the connector causally useful rather than allowing a
language model to ignore $T_E$.

\paragraph{Semantic mask calibration.}
Segmentation adds 32 learned tokens. Their final Qwen states are projected and
fused into a semantic vector $s$, which calibrates the dense field through a
bounded spatial residual:
\begin{align}
  s &= \operatorname{Fuse}(\operatorname{Proj}(H_S)),\\
  \Delta L_{ij} &= \langle F_{:,ij},s\rangle/16+b,\\
  L_{ij} &= (L_0)_{ij}+2\sigma(\alpha)\Delta L_{ij},\quad
  \mpred=\sigma(L).
\end{align}
An auxiliary head predicts anomaly presence. Dice and presence-aware binary
cross-entropy supervise the mask, with replay of understanding examples protecting the
preceding capability. DINO therefore remains the local correspondence engine while
Qwen supplies semantic context without autoregressively decoding mask pixels.
The bounded residual is the key division of labor: language may suppress an
implausible response or emphasize a semantically consistent region, but it
cannot replace the high-resolution evidence field with a free-form mask.

\subsection{Instruction-Conditioned Diffusion Editing}

Generation deliberately excludes $T_E$: it follows the edit instruction rather than
a predicted anomaly. Qwen receives the source image, instruction, and
256 learned Diffusion Queries. An instruction resampler lets these queries read the
Qwen sequence, and the encoder projection module (Enc-Proj) maps their final
states to the UNet's
cross-attention space. Following learned-query transfer in MetaQuery
\citep{pan2025metaquery}, the result
provides the diffusion model's conditional input:
\begin{equation}
  C_G=\operatorname{EncProj}(H_{\mathrm{DQ}})
  \in\R^{256\times1024}.
\end{equation}

The editor retains SD2's original nine-channel input: noisy latents, a latent edit
mask, and masked-source latents. An ROI-normalized loss emphasizes the edit
region. The source specifies what to
preserve, $\medit$ where change is allowed, and the instruction what to create.
Unlike Evidence Tokens, the 256 Diffusion Queries describe no existing anomaly:
they form the complete cross-attention condition, while the original inpainting inputs
retain pixel-aligned preservation and mask information.
This separation allows a normal source to receive a requested defect without
first predicting one, and prevents observation-side evidence from overriding
the user-specified editing region.

\subsection{Staged Capability Optimization}

Training follows five stages: dense-evidence initialization, evidence-grounded
language alignment, semantic mask calibration with replay of understanding
examples, reconstruction pretraining, and industrial edit specialization.
Restricted updates and replay preserve earlier capabilities, yielding one fixed
parameter set without task-specific checkpoints or unconstrained joint
fine-tuning. The order establishes localization before language calibration and
gives diffusion conditioning a rendering prior before industrial editing.

\begin{table*}[t]
\centering
\small
\setlength{\tabcolsep}{3.5pt}
\begin{tabular}{@{}lrrrrrrrr@{}}
\toprule
Method & AD & DefCls & DefLoc & DefDesc & DefAna & ObjCls & ObjAna & Macro$_7$ \\
\midrule
\multicolumn{9}{l}{\itshape Unified evaluation protocol} \\
Qwen3.5-9B & 70.88 & 56.94 & 62.51 & 63.17 & 74.88 & 92.31 & 78.86 & 71.36 \\
InternVL3.5-8B & 67.32 & 55.23 & 60.48 & 61.41 & 73.16 & 91.20 & 77.42 & 69.46 \\
Ovis-U1-3B & 65.51 & 53.64 & 58.29 & 59.52 & 71.84 & 90.45 & 75.83 & 67.87 \\
Gemma3-4B & 57.91 & 37.42 & 36.16 & 46.83 & 76.88 & 84.67 & 74.12 & 59.14 \\
SPHINX-7B & 53.62 & 34.17 & 51.78 & 51.44 & 71.86 & 84.69 & 73.58 & 60.16 \\
LLaVA-NeXT-34B & 58.46 & 49.12 & 53.74 & \textbf{70.88} & \textbf{80.91} & 81.36 & 78.22 & 67.53 \\
AnomalyGPT & 64.18 & 32.44 & 25.61 & 39.73 & 35.86 & 34.12 & 28.74 & 37.24 \\
\textbf{\method} & \textbf{73.47} & \textbf{58.72} & \textbf{64.36} & 64.89 & 76.60 & \textbf{93.01} & \textbf{80.13} & \textbf{73.02} \\
\bottomrule
\end{tabular}
\caption{MMAD one-shot results with equal-weight averaging across datasets. All
methods use one evaluation protocol; bold marks the column best. DefCls,
DefLoc, DefDesc, and DefAna denote defect classification, localization,
description, and analysis; ObjCls and ObjAna denote object classification and
analysis. AD is balanced accuracy; all other columns are accuracy.}
\label{tab:mmad}
\end{table*}

\section{Experiments}
\label{sec:experiments}

We evaluate cross-task competitiveness, the contribution of the token interfaces, and
the specialist tradeoffs that remain under unification.

\subsection{Experimental Protocol}

\paragraph{Understanding.}
MMAD uses a fixed one-shot protocol with 39,670 questions over 8,366 images
\citep{jiang2024mmad}. Macro$_7$ averages five defect and two object
capabilities. All listed models, including Gemma3 \citep{gemmateam2025gemma3},
SPHINX \citep{lin2023sphinx}, and LLaVA-NeXT \citep{liu2024llavanext}, use this
protocol; test questions and answers are excluded from training and selection.

\paragraph{Segmentation.}
The internal test contains 14,844 anomalous and 14,744 normal samples; metrics
are Dice, IoU, pixel AP/AUROC, and normal-pixel FPR (N-FPR). Public evaluation
uses 4,345 MVTec AD, VisA, and MPDD images with category-macro AP, AUROC, and
AUPRO-30. Baselines include DINOv2 matching \citep{oquab2023dinov2},
AnomalyDINO \citep{damm2025anomalydino}, AnomalyCLIP
\citep{zhou2026anomalyclip}, and PatchCore \citep{roth2022patchcore}; \method
uses one reference.

\paragraph{Generation.}
Controlled Pairs contains 100 edits of real normal images; Target Pairs contains
63 held-out source--target edits. We report ROI-LPIPS \citep{zhang2018lpips},
masked DINO distance, and masked PSNR. SD2 Inpainting, PowerPaint
\citep{zhuang2023powerpaint}, BrushNet \citep{ju2024brushnet}, and \method share
inputs, 30-step sampling, seeds, and binary-mask compositing.

\paragraph{Implementation.}
All \method results use one parameter set; the language backbone, visual
encoder, and VAE remain frozen. Inference uses constrained understanding
decoding, fixed one-reference segmentation, and 30-step PNDM generation with
classifier-free guidance 2.0. Test partitions select neither checkpoints nor
task-specific model variants.

\subsection{Evidence for Selective Unification}

\begin{figure}[t]
  \centering
  \includegraphics[width=\columnwidth]{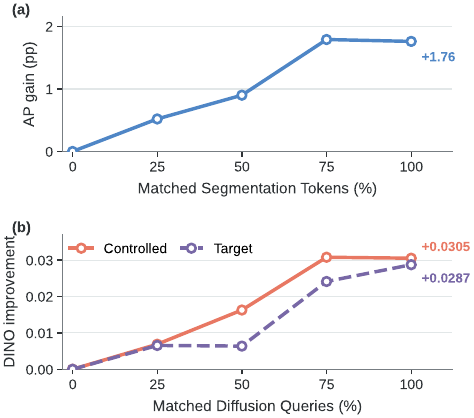}
  \caption{Replacing nuisance-matched shuffled representations with
  sample-aligned counterparts yields near-ordered, saturating gains under fixed final weights and
  test inputs. At full alignment, pixel AP improves by 1.76 points, while
  masked DINO distance decreases by 0.0305 and 0.0287 on Controlled and Target
  Pairs, respectively.}
  \label{fig:selective-unification-evidence}
\end{figure}

We vary the fraction of sample-aligned interface representations while holding
the final weights and test inputs fixed; the remaining representations are
replaced by nuisance-matched shuffled counterparts.
Figure~\ref{fig:selective-unification-evidence}
shows near-ordered gains followed by saturation in both pixel AP and masked
DINO fidelity. This response directly validates that the performance gains arise
from sample-aligned interface content rather than interface activation alone.
The common 75--100\% saturation region across segmentation and both generation
sets further indicates that the effect is shared across pixel ranking and
semantic editing rather than being specific to one metric.

\subsection{Understanding Results}

Table~\ref{tab:mmad} compares open MLLMs from 3B to 34B parameters under the
same MMAD protocol. \method reaches the highest Macro$_7$ at 73.02\%, exceeding
Qwen3.5-9B by 1.66 points and LLaVA-NeXT-34B by 5.49 points despite using a 9B
language backbone. The gain is broad rather than concentrated in one category:
\method leads anomaly detection, defect classification, defect localization,
object classification, and object analysis. LLaVA-NeXT-34B leads defect
description and analysis, while \method's five-category advantage establishes
the stronger overall profile. This same model also supports the dense and
editing evaluations below.

\subsection{Segmentation Results}

\begin{table}[t]
\centering
\small
\setlength{\tabcolsep}{3pt}
\begin{tabular}{@{}lrrrrr@{}}
\toprule
Variant & Dice & IoU & AP & AUROC & N-FPR \\
\midrule
Dense baseline & 72.52 & 61.07 & 72.90 & 91.46 & 0.0831 \\
Without aligned tokens & 72.52 & 61.07 & 72.92 & 91.50 & 0.0845 \\
\textbf{\method} & \textbf{72.66} & \textbf{61.23} & \textbf{83.77} & \textbf{95.63} & \textbf{0.0140} \\
\bottomrule
\end{tabular}
\caption{Internal segmentation on \dataset under a fixed 29,588-sample
evaluation set and protocol. Token suppression removes the aligned Evidence Tokens. Higher is
better except for N-FPR.}
\label{tab:segmentation-internal}
\end{table}

Table~\ref{tab:segmentation-internal} isolates the evidence interface. \method
improves the dense baseline by 10.87 AP and 4.17 AUROC points, reducing N-FPR
from 0.0831\% to 0.0140\%, while Dice and IoU change little. Removing aligned
tokens returns nearly to the dense baseline. This contrast shows that language
recalibrates pixel confidence rather than redrawing anomaly support, while the
lower N-FPR confirms suppression of normal-region responses.

\begin{table}[t]
\centering
\small
\setlength{\tabcolsep}{1.4pt}
\begin{tabular}{@{}lrrrrrr@{}}
\toprule
Method & MVTec & VisA & MPDD & AP & AUROC & AUPRO \\
\midrule
DINOv2-NN-1 & 47.54 & 28.73 & 30.62 & 35.63 & 93.18 & 84.86 \\
AnomalyDINO-16 & 57.58 & 42.35 & 36.38 & 45.44 & 96.74 & 91.42 \\
AnomalyDINO-1 & 51.55 & 37.48 & 32.10 & 40.38 & 95.18 & 87.76 \\
AnomalyCLIP-0 & 34.42 & 21.41 & 27.87 & 27.90 & 94.16 & 86.99 \\
PatchCore & 47.83 & 33.69 & 37.17 & 39.56 & 97.52 & 87.56 \\
\textbf{\method} & \textbf{81.18} & \textbf{44.94} & \textbf{45.18} & \textbf{57.10} & 92.55 & 77.77 \\
\bottomrule
\end{tabular}
\caption{Public segmentation. Suffixes denote reference count (0: zero-shot);
PatchCore uses its normal bank and \method uses one reference. The final columns
average category-macro results across the three datasets.}
\label{tab:segmentation-public}
\end{table}

\begin{figure}[!t]
\centering
\includegraphics[width=0.84\columnwidth]{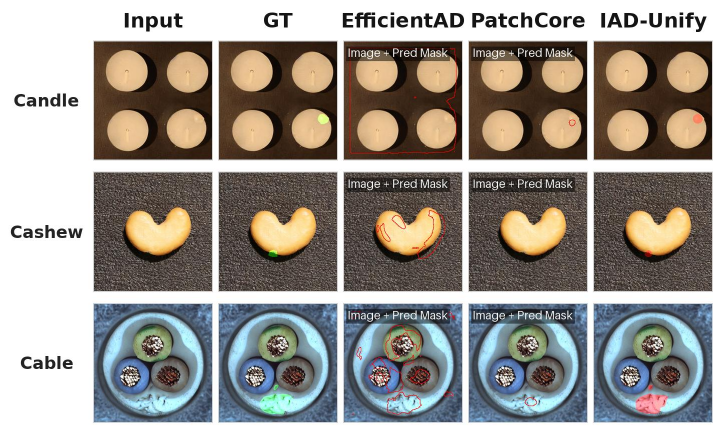}
\caption{Segmentation of contamination, missing-part, and crack anomalies on
held-out industrial images.}
\label{fig:qualitative-segmentation}
\end{figure}

Table~\ref{tab:segmentation-public} reports the highest listed AP averaged across
datasets for \method (57.10\%) using one reference, with a lead on MVTec AD,
VisA, and MPDD. Figure~\ref{fig:qualitative-segmentation} shows compact defects
with fewer background responses across three anomaly types.

\subsection{Controlled Generation Results}

\begin{figure}[!t]
\centering
\includegraphics[width=0.84\columnwidth]{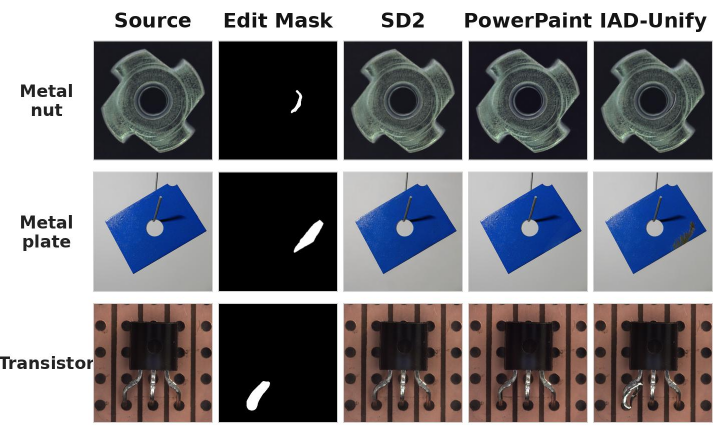}
\caption{Matched mask-guided editing of real normal sources.}
\label{fig:qualitative-generation}
\end{figure}

\begin{table}[t]
\centering
\small
\setlength{\tabcolsep}{4pt}
\begin{tabular}{@{}lrrr@{}}
\toprule
\multicolumn{4}{c}{Controlled Pairs} \\
\cmidrule(lr){1-4}
Method & LPIPS & DINO & PSNR \\
\midrule
SD2 Inpaint & 0.0205 & 0.5675 & 13.99 \\
PowerPaint v2 & \textbf{0.0195} & 0.5580 & \textbf{14.60} \\
BrushNet & 0.0257 & 0.4431 & 11.74 \\
\textbf{\method} & 0.0217 & \textbf{0.3765} & 11.60 \\
\midrule
\multicolumn{4}{c}{Target Pairs} \\
\cmidrule(lr){1-4}
Method & LPIPS & DINO & PSNR \\
\midrule
SD2 Inpaint & 0.0125 & 0.4231 & 15.56 \\
PowerPaint v2 & \textbf{0.0115} & \textbf{0.3890} & \textbf{16.86} \\
BrushNet & 0.0201 & 0.5416 & 13.01 \\
\method & 0.0133 & 0.3970 & 14.17 \\
\bottomrule
\end{tabular}
\caption{Generation under matched source, mask, instruction, sampling, and
compositing. Lower LPIPS/DINO and higher PSNR are better.}
\label{tab:generation}
\end{table}

Table~\ref{tab:generation} shows that on Controlled Pairs, \method has the lowest masked DINO distance (0.3765), 0.0666
below the closest baseline, BrushNet. PowerPaint leads LPIPS and PSNR, exposing
a semantic-edit versus pixel-reconstruction tradeoff. On Target Pairs,
PowerPaint leads, while \method remains close in DINO distance (0.3970 versus
0.3890) and outperforms BrushNet. Figure~\ref{fig:selective-unification-evidence}
further shows that matched interface content improves generation fidelity.
Thus, semantic fidelity remains competitive on held-out edits.

Figure~\ref{fig:qualitative-generation} illustrates the same-source comparison:
SD2 may under-edit, whereas \method produces clearer localized changes while
preserving unmasked context. Together, the evaluations establish competitive
reasoning, evidence-calibrated pixel ranking, and semantically aligned editing
in one model while preserving task-specific inputs.

\paragraph{Cross-task synthesis.}
Evidence Tokens calibrate dense evidence, while Diffusion Queries modify
cross-attention without replacing the source or mask. Figure~\ref{fig:selective-unification-evidence}
confirms gains from aligned content under fixed parameters and inputs. Together
with MMAD, this distinguishes selective unification from a common API: shared
semantics and parameters improve all three tasks without one lossy bottleneck.

\section{Conclusion}
\label{sec:conclusion}

\method combines industrial anomaly understanding, segmentation, and localized
generation through task-specific token interfaces. One parameter set reaches
73.02\% MMAD Macro$_7$, 57.10\% public segmentation AP, and the best Controlled
masked DINO result. Dense correspondence, grounded tokens, and source-mask
conditioning remain tied to localization, reasoning, and editing, respectively.
The results show that selective unification can share semantic evidence without
collapsing spatial reasoning and image synthesis into a single representation,
retaining competitive task-specific behavior within one model. \dataset
provides family-disjoint supervision for this design. This turns unification
into an explicit routing problem rather than a compromise among incompatible
outputs; future work will extend it beyond mask-guided industrial editing.

\bibliography{references}

\end{document}